\documentclass[11pt,amsfonts]{article}
\usepackage{amssymb}
\usepackage{epstopdf}
\usepackage{shuffle}
\usepackage{tikz}
\usepackage{graphicx}
\usepackage[hyphens]{url}
\usepackage{microtype}
\usepackage{url}
\def\P{{\mathchoice {\hbox{$\sf\textstyle P\kern-0.4em Z$}}
{\hbox{$\sf\textstyle P\kern-0.4em P$}}
{\hbox{$\sf\scriptstyle P\kern-0.3em P$}}
{\hbox{$\sf\scriptscriptstyle P\kern-0.2em P$}}}}

\usepackage{epsf}
\def\P{{\mathchoice {\hbox{$\sf\textstyle P\kern-0.4em Z$}}
{\hbox{$\sf\textstyle P\kern-0.4em P$}}
{\hbox{$\sf\scriptstyle P\kern-0.3em P$}}
{\hbox{$\sf\scriptscriptstyle P\kern-0.2em P$}}}}

\let\Rightarrow=\Rightarrow
\newtheorem{theorem}{\bf Theorem}[section] 
{}
\newtheorem{prob}{\bf Problem}[section]
\newtheorem{proposition}[theorem]{\bf Proposition} 
\newtheorem{corollary}[theorem]{\bf Corollary}
\newtheorem{lemma}[theorem]{\bf Lemma} \newtheorem{defin}[theorem]{\bf
Definition}
\newtheorem{rema}[theorem]{\bf Remark}
 \newtheorem{example}[theorem]{\bf Example}
\newtheorem{condit}{\bf Condition}
\newtheorem{fait}{\bf Claim}
\newtheorem{notat}{\bf Notation}

\def\ex#1\par{\par\noindent\begin{exemple} \nopagebreak \strut \rm #1 \end{exemple}}
\def\thm#1\par{\medskip\par\noindent\begin{theorem} \strut \sl #1 \end{theorem}\par}
\def\propo#1\par{\medskip\par\noindent\begin{proposition} \strut \sl #1 \end{proposition} \par}
\def\proo#1\par{\medskip\par\noindent{\it Proof.} \strut \rm #1 $\Box$ \par}
\def\prob#1\par{\medskip\par\noindent\begin{prob} \strut \sl #1 \end{prob} \par}
\def\cor#1\par{\medskip\par\noindent\begin{corollary} \strut \sl #1 \end{corollary}\par}
\def\lm#1\par{\medskip\par\noindent\begin{lemma} \strut \sl #1 \end{lemma}\par}
\def\defil#1\par{\medskip\par\noindent\begin{condit} \strut \sl #1\end{condit}\par}
\def\fct#1\par{\medskip\par\noindent\begin{fait} \strut \sl #1 \end{fait}\par}
\def\defi#1\par{\medskip\par\noindent{\begin{defin} \strut  \sl #1 \end{defin}}\par}
\def\nota#1\par{\par\noindent\begin{notat} \nopagebreak  \strut #1 \end{notat}}

\def\rem#1\par{\par\noindent\begin{rema} \nopagebreak \strut \rm #1 \end{rema}}

\def\ex#1\par{\par\noindent\begin{example} \nopagebreak \strut #1 \end{example}}

\date{}
\begin{document}

\thispagestyle{empty}
\title{\bf Complete Variable-Length Codes: An Excursion into Word Edit Operations}
%
\author{Jean N\'ERAUD
\thanks{Universit\'e de Rouen, Laboratoire d'Informatique, de Traitement de l'Information et des Syst\`emes, 
 Avenue de l'Universit\'e, 76800 Saint-\'Etienne-du-Rouvray,
France; jean.neraud@univ-rouen.fr, neraud.jean@gmail.com; http:neraud.jean.free.fr; orcid: 0000-0002-9630-461X}
}

\maketitle
{\small
\begin{center}
{\bf Abstract}
\end{center}
Given  an alphabet $A$ and a binary relation $\tau\subseteq A^*\times A^*$, a language $X\subseteq A^*$
is $\tau$-{\it independent}
 if $ \tau(X)\cap X=\emptyset$; $X$  is  $\tau$-{\it closed} if $\tau(X)\subseteq X$. 
The language $X$  is {\it complete} if any word over $A$ is  a factor of some concatenation of words in $X$. 
Given a family of languages  ${\cal F}$ containing $X$, $X$  is maximal in  ${\cal F}$ if no other set  of ${\cal F}$ can stricly contain $X$.
A language $X\subseteq A^*$ is a {\it variable-length} code if any equation among the words of $X$ is necessarily trivial.
The study discusses the  relationship between maximality and completeness in the case of $\tau$-independent or $\tau$-closed variable-length codes.
We focus to the binary relations by which the images of words
are computed by deleting, inserting, or substituting some characters.
}
\maketitle
\section{Introduction}
\label{Introduction}
In formal language theory, given a property ${\cal F}$, the {\it embedding problem} with respect to ${\cal F}$  consists in examining whether a  language $X$ satisfying ${\cal F}$ can be included into some language  $\hat X$  that is {\it maximal}  with respect to ${\cal F}$,
 in the sense that no language satisfying ${\cal F}$  can strictly contain  ${\hat X}$. In the literature, maximality is often connected to completeness: 
a language  $X$ over the alphabet $A$ is {\it complete} if any string in the free monoid $A^*$ (the set of the words over $A$) is a factor of some word of $X^*$ (the submonoid of all concatenations of words in $X$). Such connection takes on special importance for codes:
a language $X$ over the alphabet $A$  is a {\it variable-length code} (for short, a {\it code}) if every equation among the {\it words} (i.e. {\it strings})
of $X$ is necessarily trivial.

A famous result due to M.P. Sch\"utzenberger states that, for  the family of the so-called {\it thin} codes (which contains {\it regular} codes and therefore also  finite ones), being maximal is equivalent to being complete.
In connection with these two concepts lots of challenging  theoretical questions have been stated.
 For instance, to this day the problem of the existence of a finite maximal code containing a given finite  one  is not known to be decidable.
 From this latter point of view, in \cite{R77} the author asked the question of the existence of a regular complete code containing a given finite one: a positive answer was brought in  \cite{ER85}, 
where was provided a now classical formula for embedding a given regular code  into some complete regular one.
Famous families of codes have also been concerned by those studies: we mention
{\it prefix} and {\it  bifix} codes \cite[Proposition 3.3.8, Proposition 6.2.1]{BPR10},
codes with a {\it finite deciphering delay}  \cite{BWZ90},
{\it infix} \cite{L00},  {\it solid} \cite{L01}, 
or {\it circular}  \cite{N08}. 

Actually, with each of those families,  a so-called {\it dependence system} can be associated. 
Formally, such a system is  a family ${\cal F}$ of languages   
constituted by those sets $X$
that contain  a non-empty finite subset  in ${\cal F}$. Languages in ${\cal F}$ are ${\cal F}$-{\it dependent}, the other ones being ${\cal F}$-{\it independent}.
A special case corresponds to 
binary words relations
$\tau\subseteq A^*\times A^*$, where a dependence systems is constituted by those sets $X$ satisfying $\tau\cap (X\times X)\neq\emptyset$:
$X$ is $\tau$-{\it independent} if we have
$\tau(X)\cap X=\emptyset$ (with $\tau(X)=\{y: \exists x\in X, (x,y)\in\tau\}$).
{\it Prefix codes}  certainly constitute  the best known example: 
they constitute those codes that are independent with respect to the relation obtained  
by removing each pair $(x,x)$ from the famous {\it prefix} order.
Bifix, infix or solid codes can be similarly characterized.

As regards to dependence, some extremal condition corresponds to the so-called {\it closed} sets: 
given a  word relation $\tau\subseteq A^*\times A^*$,  a language $X$  is closed under $\tau$ ($\tau$-{\it closed}, for short) if 
we have
$\tau(X)\subseteq X$.
Lots of topics are concerned by the notion. We mention the framework of  prefix order where a one-to-one correspondence between independent  and closed sets is provided in \cite[Proposition 3.1.3]{BPR10} (cf. also \cite{BDFPRR12,RMW90}). 
Congruences in the free monoid are also concerned \cite{Ni71}, as well as their connections to  DNA computing \cite{KPTY97}. 
With respect to morphisms, involved topics are also provided  by the famous  $L$-systems \cite{RS80} and,
in the  case of one-to-one (anti)-automorphisms,
the so-called {\it invariant} sets \cite{NS18}.

As commented in \cite{JK97}, maximality and completeness concern the economy of a code. If $X$ is a
complete code then every word occurs as part of a message, hence no part of $X^*$ is potentially useless. 
The present paper emphasizes  the following questions: 
given a regular binary relation $\tau\subseteq A^*\times A^*$,
in the family of regular  $\tau$-independent (-closed) codes, are  maximality and completeness equivalent notions?
Given a non-complete regular $\tau$-independent (-closed) code, is it embeddable into some complete one?

Independence has some peculiar importance in the framework of coding theory. Informally, given some concatenation of words in $X$, each codeword $x\in X$  is transmitted via a channel into a corresponding $y\in A^*$.
According to the combinatorial structure of $X$, and the type of channel,  one  has to make use of codes with prescribed error-detecting constraints: some minimum-distance restraint is generally applied.
 In this paper, where we consider variable length codewords, we address to the Levenshtein metric \cite{L65}:
given two different words $x,y$, their distance is the minimal total number of elementary edit operations that can transform $x$ into $y$, such operation consisting in a one character
 {\it deletion}, {\it insertion}, or {\it substitution}. Formally, it is the smallest integer $p$ such that we have $y\in\Lambda_p(x)$, with $\Lambda_p=\bigcup_{1\le k\le p}(\delta_1\cup \iota_1\cup \sigma_1)^k$,  where $\delta_k$, $\iota_k$, $\sigma_k$ are further defined below.
From the point of view of error detection, $X$ being $\Lambda_p$-independent guarantees that $y\in\Lambda_p(x)$ implies $y\neq x$.
In addition, a code satisfies the property of error correction if its elements are such that $\Lambda_p(x)\cap\Lambda_p(y)=\emptyset$ unless $x=y$: 
according to  \cite[chap. 6]{K96}, the existence of such codes is decidable.
Denote by  Subw($x$) the set of the subsequences of $x$:  

\smallskip
-- $\delta_k$, the {\it $k$-character deletion},  associates with every word $x\in A^*$, all the words $y\in {\rm Subw}(x)$  whose length is $|x|-k$. The {\it at most $p$-character deletion} is $\Delta_p=\bigcup_{1\le k\le p}\delta_k$;

-- $\iota_k$, the {\it $k$-character insertion},  is the converse relation of $\delta_k$ and we set $I_p=\bigcup_{1\le k\le p} \iota_k$ ({\it at most $p$-character insertion});

--  $\sigma_k$, the {\it $k$-character substitution},  associates with every $x\in A^*$, all $y\in A^*$ with length $|x|$  such that $y_i$ (the letter of position $i$ in $y$), differs of $x_i$ in exactly $k$ positions $i\in [1,|x|]$; we set $\Sigma_p=\bigcup_{1\le k\le p}\sigma_k$;

-- We denote by  ${\underline \Lambda}_p$ the antireflexive relation obtained by removing all pairs $(x,x)$ from $\Lambda_p$ (we have $\Lambda_1={\underline \Lambda}_1$).

\smallskip
For short, we will refer the preceding relations  to {\it edit relations}. For reasons of consistency,  in the whole paper we assume $|A|\ge 2$ and $k\ge 1$. In what follows, we draw the main contributions of the study: 

\smallskip
Firstly,
 we prove that, given a positive integer $k$, the two families of languages that are independent with respect to $\delta_k$  or $\iota_k$ are identical. In addition, 
for $k\ge 2$, no set can be $\Lambda_k$-independent. We  establish the following result:
{\flushleft {\bf Theorem~A.}}
{\it Let 
$A$ be a finite alphabet, $k\ge 1$, and $\tau\in\{\delta_k,\iota_k,\sigma_k,\Delta_k,I_k,\Sigma_k,{\underline \Lambda}_k\}$.
Given a regular $\tau$-independent code $X\subseteq A^*$, $X$ is complete if, and only if,
it is maximal in the family of $\tau$-independent codes.}

\medskip
A code $X$  is ${\underline \Lambda}_k$-independent if the Levenshtein distance between two distinct  words of $X$  is always larger than $k$: from this point of view,  Theorem A  states some noticeable characterization of maximal $k$-error detecting codes in the framework of the Levenshtein metric. 

\medskip
 Secondly, we  explore  the domain of closed codes.
A noticeable fact is that for any $k$, there are only finitely many $\delta_k$-closed codes and they have finite cardinality. 
Furthermore, one can  decide whether a given non-complete $\delta_k$-closed code can be embedded into some complete one. 
 We also  prove  that  no closed code can exist with respect to the relations $\iota_k$, $\Delta_k$, $I_k$.
 
As regard to substitutions, beforehand, we focus to the structure of the set $\sigma_k^*(w)=\bigcup_{i\in {\mathbb N}}\sigma_k^i$.
Actually, excepted for two special cases (that is, $k=1$ \cite{E73,S97}, or $k=2$ with $|A|=2$ \cite[ex. 8, p.77]{K05}), to our best knowledge, in the literature no general description is provided.
In any event we provide such a description;
furthermore we establish the following result:
{\flushleft {\bf Theorem B}.}
{\it Let $A$ be a finite alphabet and $k\ge 1$. Given  a complete $\sigma_k$-closed code $X\subseteq A^*$,
either  every word in $X$ has length not greater than $k$, or a unique integer $n\geq k+1$ exists such that $X=A^n$.
In addition for every $\Sigma_k$($\Lambda_k$)-closed code $X$, some positive integer $n$ exists such that $X=A^n$.}

\medskip
In other words, no $\sigma_k$-closed code can simultaneously possess words in $A^{\le k}$
and words in $A^{\ge k+1}$. 
As a consequence, one can decide whether a given non-complete $\sigma_k$-closed code $X\subseteq A^*$  is embeddable into some complete one.
\section{Preliminaries}
We adopt the notation of the free monoid theory.
Given a word $w$, we denote by $|w|$ its length; 
for $a\in A$,  $|w|_a$ denotes the number of occurrences of the letter $a$ in $w$.
 The set of the words whose length is not greater (not smaller) than $n$ is denoted by $A^{\le n}$ ($A^{\ge n}$).
 Given $x\in A^*$ and $w\in A^+$, we say that $x$ is a {\it factor}  of $w$ if words $u,v$ exist such that $w=uxv$; 
a {\it subword} of $w$ consists in any (perhaps empty) subsequence $w_{i_1}\cdots w_{i_n}$ of $w=w_1\cdots w_{|w|}$.
We denote by
${\rm {\rm F}(}X)$ (${\rm Subw}(X)$) the set of the words that are factor (subword) of some word in $X$ (we have $X\subseteq {\rm {\rm F}(}X)\subseteq {\rm Subw}(X)$). 
A pair of words $w,w'$ is {\it overlapping-free} if no pair  $u,v$ exist such that $uw'=wv$ or $w'u=vw$,
with $1\le |u| \le |w|-1$ and  $1\le |v| \le |w'|-1$; 
if $w=w'$, we say that $w$ itself is overlapping-free.

 It is assumed that the reader has a fundamental understanding  with the main concepts of the theory of variable-length codes: we 
suggest, if necessary,  that  he (she) report to \cite{BPR10}.
A set $X$ is a {\it variable-length code} (a {\it code} for short) if  for any pair of sequences of words in $X$, say  $(x_i)_{1\le i\le n}$, $(y_j)_{1\le j\le p}$, the equation
$x_1\cdots x_n=y_1\cdots y_p$ implies  $n=p$, and $x_i=y_i$ for each integer $i$ (equivalently the submonoid $X^*$ is {\it free}).
The two following results are famous ones from  the variable length code theory:
\begin{theorem}{\rm Sch\"utzenberger \cite[Theorem 2.5.16]{BPR10}}
\label{classic}
Let $X\subseteq A^*$ be a regular code.
Then the following properties are equivalent:

{\rm (i)} $X$ is complete;

{\rm (ii)} $X$ is a maximal code;

{\rm (iii)} a positive Bernoulli distribution $\pi$ exists such that $\pi(X)=1$;

{\rm (iv)} for every positive Bernoulli distribution $\pi$ we have $\pi(X)=1$.
\end{theorem}
\begin{theorem}{\rm  \cite{ER85}}
\label{EhrRoz}
Given a non-complete code $X$, let $y\in A^*\setminus {\rm F}(X^*)$ be an overlapping-free word and  $U=A^*\setminus (X^*\cup A^*yA^*)$. Then $Y=X\cup y(Uy)^*$ is a complete code.
\end{theorem}
With regard to word relations, the following statement comes from the definitions:
\begin{lemma}
\label{iteration-tau-sur-X}
Let $\tau\in A^*\times A^*$ and $X\subseteq A^*$. Each of the following properties holds:

{\rm (i)} $X$ is  $\tau$-independent if, and only if, it is  $\tau^{-1}$-independent ($\tau^{-1}$ denotes the converse relation of $\tau$).

{\rm (ii)} $X$ is $\delta_k$($\Delta_k$)-independent if, and only if, it is $\iota_k$($I_k)$-independent.

{\rm (iii)} $X$ is $\tau$-closed if, and only if, it is $\tau^*$-closed.
\end{lemma}
\section{Complete independent codes}
\label{independent-dis}
 We start by providing a few  examples:
\begin{example}
\label{EYYY}
 For  $A=\{a,b\}$, $k=1$, the prefix code $X=a^*b$ is not $\delta_k$-independent (we have $a^{n-1}b\in\delta_k(a^nb)$), whereas the following codes are $\delta_1$-independent:

-- the regular (prefix) code: $Y=\{a^{2}\}^+\{b,aba,abb\}$.
Note that since it contains $\{a^{2}\}^+$,  $\delta_1(Y)$ is not a code.

-- the complete (non-regular) context-free Dyck bifix code  $D_1$, which generates  the Dyck free submonoid $D_1^*=\{w\in A^*: |w|_a=|w_b|\}$ (for every word $w\in D_1$ we have $|\delta_1(w)|_a\neq |\delta_1(w)|_b$). Note that $\delta_1(D_1)$ contains  the empty word, $\varepsilon$,   thus it cannot be a code; however $\delta_1(D_1)\setminus A$ remains a (non-complete) bifix code

-- the non-complete finite bifix code $Z=\{ab^2,ba^2\}$: actually, $\delta_1(Z)$ is the complete uniform code $A^2$.

-- for every pair of different integers $n,p\ge 2$, the prefix code $T=aA^n\cup bA^p$. We have $\delta_1(T)=A^n\cup A^p$, which  is not a code, although it is complete.
\end{example}
In view of establishing the main result of Section \ref{independent-dis}, we will construct some peculiar word:
\begin{lemma}
\label{incompletable-del-ins}
Let $k\ge1$, $i\in [1,k]$, $\tau\in\{\delta_i,\iota_i,\sigma_i\}$.
Given a   a non-complete
 code $X\subseteq A^*$ some overlapping-free word $y\in A^*\setminus {\rm F}(X^*)$ exists such that $\tau(y)$ does not intersect $X$  and $y\notin\tau(X)$.
\end{lemma}
{\it Proof.}
Let $X$ be a non-complete code, and let $w\in A^*\setminus {\rm {\rm F}(}X^*)$. Trivially, we have $w^{k+1}\notin {\rm {\rm F}(}X^*)$.
Moreover, in a classical way a word $u\in A^*$ exists such that $y=w^{k+1}u$ is overlapping-free (eg. \cite[Proposition 1.3.6]{BPR10}).
Since we assume $i\in [1,k]$, each word in $\tau(y)$ is constructed by  deleting (inserting, substituting)  at most  $k$ letters from  $y$, hence by construction it contains at least one occurrence of $w$ as a factor.
This implies  $\tau(y)\cap{\rm {\rm F}(}X^*)=\emptyset$,
thus  $\tau(y)$ does not intersect $X$.

By contradiction, assume that a word $x\in X$ exists such that $y\in\tau(x)$.
It follows from $\delta_k^{-1}=\iota_k$ and $\sigma_k^{-1}=\sigma_k$ that  $y=w^{k+1}u$ is obtained by deleting (inserting, substituting)  at most
 $k$ letters from $x$: consequently at least one occurrence of $w$ appears as a factor of $x\in X\subseteq {\rm F(}X^*{\rm )}$: this contradicts $w\notin {\rm {\rm F}(}X^*)$, therefore we obtain  $y\notin\tau(X)$ (cf. Figure 1).
$\Box$\\

\bigskip
%
\includegraphics[width=12cm,height=2.75cm]{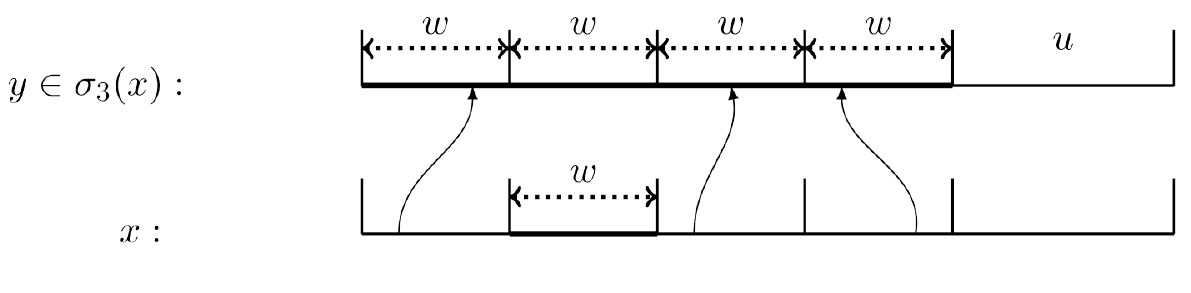}
\ \\
\begin{center}
Fig. 1: {\footnotesize Proof of  Lemma  \ref{incompletable-del-ins}: $y\in\tau(X)$ implies $w\in {\rm F}(X)$;  for  $i=k=3$ and $y=w^4u$,  the action of  the substitution $\tau=\sigma_3$ is represented by the arrows, in some extremal condition.}
\end{center}
%
As a consequence, we obtain the following result:
\begin{theorem}
\label{classic4}
Let  $k\ge 1$ and  $\tau\in\{\delta_k,\iota_k,\sigma_k\}$.
Given a regular $\tau$-independent code $X\subseteq A^*$,  $X$ is complete if, and only if,
it is maximal as an $\tau$-independent codes.
\end{theorem}
{\it Proof.}
According to Theorem \ref{classic}, every complete $\tau$-independent code is a maximal code, hence it is maximal in the family of $\tau$-independent codes. 
For proving the converse, we make use of the contrapositive.
Let $X$ be a non-complete $\tau$-independent code, and let $y\in A^*\setminus {\rm {\rm F}(}X^*)$ satisfying the conditions of Lemma \ref{incompletable-del-ins}.
With the notation of Theorem \ref{EhrRoz},
necessarily  $X\cup\{y\}$, which is a subset of  $Y=X\cup y(Uy)^*$, is a code. 
According to Lemma \ref{incompletable-del-ins}, we have  $\tau(y)\cap X=\tau(X)\cap \{y\}=\emptyset$. Since  $X$ is $\tau$-independent and $\tau$ antireflexive,
 this implies $\tau(X\cup\{y\})\cap (X\cup\{y\})=\emptyset$, thus $X$ non-maximal as a $\tau$-independent code.
$\Box$\\~\ \\
We notice that for $k\ge 2$ no $\Lambda_k$-independent set can exist
(indeed, we have $x\in\sigma_1^2(x)\subseteq \Lambda_k(x) $).
However,  the following result holds:
\begin{corollary}
\label{coro1}
Let  $\tau\in\{\Delta_k,I_k,\Sigma_k, {\underline \Lambda}_k\}$.
Given a regular $\tau$-independent code $X\subseteq A^*$, $X$ is complete if, and only if, it is maximal  as a $\tau$-independent code.
\end{corollary}
{\it Proof.} 
As indicated above, if $X$ is complete, it is  maximal as a $\tau$-independent code.
For the converse, once more we argue by contrapositive that is, with the notation  of Lemma \ref{incompletable-del-ins},
we prove that $X\cup\{y\}$ remains independent. 
By definition, for each $\tau\in\{\Delta_k,I_k,\Sigma_k,{\underline\Lambda}_k\}$, we have $\tau\subseteq\bigcup_{1\le i\le k}\tau_i$, with $\tau_i\in\{\delta_i,\iota_i,\sigma_i\}$.
According to Lemma \ref{incompletable-del-ins},  since $\tau_i$ is antireflexive, for each $i\in [1,k]$ we have $\tau_i(X\cup\{y\})\cap (X\cup \{y\})=\emptyset$:
this implies $(X\cup\{y\})\cap\bigcup_{1\le i\le k}\tau_i(X\cup\{y\})=\emptyset$, thus $X\cup\{y\}$ $\tau$-independent.
$\Box$\\~\ \\
With regard to the relation ${\underline \Lambda}_k$, Corollary \ref{coro1} expresses some interesting property in term of error detection. Indeed, as indicated in Section \ref{Introduction}, every code is ${\underline \Lambda}_k$-independent if the Levenshtein distance between its (distinct) elements is always larger than $k$. 
From this point, Corollary \ref{coro1} states some characterization of the maximality in the family of such codes.

\medskip
It should remain to develop some method in view of embedding a given non-complete  ${\underline \Lambda}_k$-code into a complete one. Since the construction from the proof  Theorem \ref{EhrRoz} does not preserve independence, this question remains open.
\section{Complete closed codes with respect to deletion or insertion}
\label{deletion-insertion}
We start with relation the $\delta_k$. 
A noticeable fact is that corresponding closed  codes are necessarily finite, as attested by the following result:
\begin{proposition}
\label{complete-delta-closed}
Given  a $\delta_k$-closed code $X$, and 
$x\in X$, we have $|x|\in [1,k^2-k-1]\setminus\{k\}$.
\end{proposition}
{\it Proof.}
It follows from $\varepsilon\notin X$ and $X$ being $\delta_k$-closed that $|x|\neq k$.
By contradiction, assume $|x|\ge (k-1)k$ and
let $q,r$ be the unique pair of integers such that $|x|=qk+r$, with $0\le r\le k-1$. 
Since we have $0\le rk\le (k-1)k\le |x|$,  an integer $s\geq 0$ exists such that $|x|=rk+s$,
thus words $x_1,\cdots, x_k,y$ exist such that 
$x=x_1\cdots x_ky$, with  
$|x_1|=\cdots=|x_k|=r$ and $|y|=s$.
By construction, every word $t\in{\rm Sub}(x)$ with  $|t|\in\{r,s\}$ belongs to $\delta_k^*(x)\subseteq X$
(indeed, we have $r=|x|-qk$ and $s=|x|-rk$).
This implies  $x_ 1,\cdots, x_k,y\in X$, thus $x\in X^{k+1}\cap X$: a contradiction with $X$ being a code.
$\Box$
\begin{example}
\label{EEXX}
 (1) According to Proposition \ref{complete-delta-closed}, no code can be $\delta_1$-closed. This can be also drawn from the fact that, for every set $X\subseteq A^+$ we have $\varepsilon\in\delta_1^*(X)$.

(2) Let $A=\{a,b\}$ and $k=3$. According to Proposition \ref{complete-delta-closed}, every word in any $\delta_k$-closed code has length not greater than $5$. 
It is straightforward to verify that  $X=\{a^2, ab, b^2,  a^4b, ab^4\}$ is a $\delta_k$-closed code.
In addition, a finite number of examinations leads to verify that $X$ is maximal  as a $\delta_k$-closed code.
Taking for $\pi$ the uniform distribution we have $\pi(X)=3/4+1/16<1$: thus  $X$ is non-complete.
\end{example}
According to Example \ref{EEXX} (2), no result similar to Theorem \ref{classic4} can be stated in the framework of $\delta_k$-closed codes. 
We also notice that, in Proposition \ref{complete-delta-closed} the bound does not depend of the size of the alphabet, but only depends of $k$.
\begin{corollary}
Given a finite alphabet $A$ and  a positive integer $k$,
one can  decide whether a non-complete $\delta_k$-closed code $X\subseteq A^*$ is included into some complete one.
In addition there are a finite number of such complete codes, all of them being computable, if any.
\end{corollary}
{\it Proof.}
According to Proposition \ref{complete-delta-closed} only a finite number of $\delta_k$-closed codes over $A$ can exist,
each of them being a subset of $A^{\le k^2-k-1}\setminus A^k$.
$\Box$\\~\ \\
We close the section by considering  the relations $\Delta_k$, $\iota_k$ and $I_k$:
\begin{proposition}
\label{no-closed-code}
No code can be $\iota_k$-closed, $\Delta_k$-closed,  nor $I_k$-closed.
\end{proposition}
{\it Proof.}
By contradiction assume that some $\iota_k$-closed code $X\subseteq A^*$ exists.
Let $x\in X$, $n=|x|$ and $u,v\in A^*$ such that $x=uv$. It follows from $|(vu)^k|=kn$,
that $u(vu)^kv\in \iota_k^*(x)$.
According  to Lemma \ref{iteration-tau-sur-X}(iii), we have  $\iota_k^*(X)\subseteq X$, thus $u(vu)^kv\in X$.
Since  $u(vu)^kv=(uv)^{k+1}=x^{k+1}\in X^{k+1}$, we have $X^{k+1}\cap X\neq\emptyset$:
a contradiction with $X$ being a code.  Consequently no $I_k$-closed codes can exist.
According to Example \ref{EEXX}(1), given a code $X\subseteq A^*$, we have $\delta_1(X)\not\subseteq X$: 
this implies $\Delta_k(X)\not\subseteq X$, thus $X$ not $\Delta_k$-closed.
$\Box$
\section{Complete codes closed under substitutions}
\label{BPR}
Beforehand,  given a word $w\in A^+$, we need a  thorough description of  the set $\sigma_k^*(w)$.
Actually, it is well known  that, over a binary alphabet, all $n$-bit words can be computed by making use of some Gray sequence  \cite{E73}. With our notation, we have $A^n=\sigma_1^*(w)$.
Furthermore, for every finite alphabet $A$, the so-called $|A|$-arity Gray sequences allow to generate  $A^n$ \cite{K05,S97}: once more we have $\sigma_1^*(w)=A^n$.
In addition, in the special case where  $k=2$ and $|A|=2$, it can be proved that we have $|\sigma_2(w)|=2^{n-1}$  \cite[Exercise 8, p. 28]{K05}.
However,  except in  these special cases, to the best of our knowledge no general description of the structure of $\sigma_k^*(w)$ appears in the literature. 
In any event, in the next paragraph we provide an exhaustive description of $\sigma_k(w)$.
Strictly speaking, the  proofs,  that we have reported in Section \ref{Proofs}, are not involved in $\sigma_k$-closed codes: we suggest the reader that,  in  a first reading, after  para. \ref{Basic}
he (she) directly  jumps  to para. \ref{Cons}.
\subsection{Basic results concerning $\sigma_k^*(w)$}
\label{Basic}
\begin{proposition}
\label{A>2-card-Sk*}
Assume $|A|\ge 3$. For each  $w\in A^{\ge k}$, we have $\sigma_k^*(w)=A^{|w|}$.
\end{proposition}
In the case where $A$ is a binary alphabet, we set $A=\{0,1\}$: this allows a well-known algebraic interpretation of $\sigma_k$.
Indeed, denote by $\oplus$ the addition in the group ${\mathbb Z}/2{\mathbb Z}$ with identity $0$,
and  fix a positive integer $n$; given $w,w'\in A^n$, define $w\oplus w'$ as the unique word of $A^n$ such that, for each $i\in [1,n]$, the letter of position $i$ in $w\oplus w'$ is $w_i\oplus w'_i$. 
With this notation the sets $A^n$ and $({\mathbb Z}/2{\mathbb Z)}^n$ are in one-to-one correspondence.
Classically, we have $w'\in\sigma_1(w)$ if, and only if, some $u\in A^n$ exists such that  $w'=w\oplus u$ with  $|u|_1=1$ (thus $|u|_0=n-1$).
From the fact that $\sigma_k(w)\subseteq\sigma_1^k(w)$,  the following property holds:
\begin{eqnarray}
\label{P}
w'\in\sigma_k(w)\Longleftrightarrow \exists u\in A^n: w=w'\oplus u,~~|u|_1=k.  
 \end{eqnarray}
In addition $w\oplus u=w'$ is equivalent to $u=w\oplus w'$.
Let $d=|\{i\in [1,n]: w_i=w'_i=1\}|$. The following property follows from 
$|u|_1=(|w|_1-d)+(|w'|_1-d)$ and $ |w|_1+|w'|_1=|w_1|+|w'|_1-2|w'|_1 \pmod{2}$
:
\begin{eqnarray}
\label{Q}
 |w|_1+|w'|_1=|w_1|-|w'|_1\pmod{2}=|u|_1\pmod{2}.
 \end{eqnarray}
Finally, for $a\in A$ we denote by ${\overline a}$ its complementary letter that is, ${\overline a}=a\oplus 1$;
for $w\in A^n$ we set ${\overline w}={\overline w_1}\cdots {\overline w_n}$.
\begin{lemma}
\label{C}
Let $A=\{0,1\}$, $n\ge k+1$. Given $w,w'\in A^n$ the two following properties hold:

{\rm (i)} If $k$ is even and $w'\in\sigma_k^*(w)$ then $|w'|_1-|w|_1$ is an even integer;

{\rm (ii)} If $|w'|_1-|w|_1$ is even then we have $w'\in\sigma_k^*(w)$, for every $k\ge 1$.
\end{lemma}
Given a positive integer $n$, we denote $A_0^n$ ($A_1^n$ ) the set of the words $w\in A^n$ such that $|w|_1$ is even (odd).
\begin{proposition}
\label{card-Sk*}
Assume $|A|=2$. Given  $w\in A^{\ge k}$ exactly one of  the following conditions holds:

{\rm (i)}  $|w|\ge k+1$,  $k$ is even, and $\sigma_k^*(w)\in\{A_0^{|w|},A_1^{|w|}\}$;

{\rm (ii)} $|w|\ge k+1$, $k$ is odd, and  $\sigma_k^*(w)=A^{|w|}$;

{\rm (iii)} $|w|=k$ and $\sigma_k^*(w)=\{w,{\overline w}\}$.
\end{proposition}
\subsection{Proofs of the statements \ref{A>2-card-Sk*}, \ref{C} and \ref{card-Sk*}}
\label{Proofs}
Actually, Proposition \ref{A>2-card-Sk*} is a consequence of the following property:
\begin{lemma}
\label{A}
Assume  $|A|\ge 3$. For every word $w\in A^{\ge  k}$ 
we have $\sigma_1(w)\subseteq \sigma_k^2(w)$.
\end{lemma}
{\it Proof.}
Let $w'\in\sigma_1(w)$ and $n=|w|=|w'|\ge k$. We prove that $w''\in A^*$ exists with
$w''\in \sigma_k(w)$ and $w'\in\sigma_k(w'')$.
By construction, $i_0\in [1,n]$ exists such that:

(a)  $w'_{i}=w_{i}$ if, and only if, $i\neq i_0$.\\
It follows from $k\le n$ that some  $(k-1)$-element subset $I\subseteq [1,n]\setminus \{i_0\}$ exists.
Since we have  $|A|\ge 3$, some letter $c\in A\setminus\{w_{i_0},w'_{i_0}\}$ exists.
Let  $w''\in A^n$ such that:

(b) $w''_{i_0}=c$ and, for each $i\neq i_0$: $w''_i\neq w_i$ if, and only if, $ i\in I$.\\
By construction we have $w''\in\sigma_k(w)$, moreover  $c\neq w'_{i_0}$ implies  $w'_{i_0}\neq w''_{i_0}$.
According to (a) and (b),
we obtain:

(c) $w'_{i_0}\neq c=w''_{i_0}$,

(d) $w'_i=w_i\neq w''_i$ if $i\in I$,  and:

(e)  $w'_i=w_i=w''_i$ if $i\notin I\cup\{i_0\}$.\\
Since we have $|I\cup\{i_0\}|=k$, this implies $w'\in\sigma_k(w'')$. 
$\Box$\\~\ \\
{\it Proof of Proposition \ref{A>2-card-Sk*}.}
Let $w'\in A^n\setminus\{w\}$: we prove that $w'\in\sigma_k^*(w)$.
 Let
$I=\{i_0,\cdots,i_p\}=\{i\in [1,n]: w'_{i}\neq w_i\}$ and
let $(w^{(i_j)})_{0\le j\le p}$ be a sequence of words such that $w=w^{(i_0)}$, $w^{(i_p)}=w'$ and, for each $j\in [0,p-1]$:
$w^{({i_{j+1}})}_\ell\neq w^{({i_{j}})}_\ell$ if, and only if, $\ell=i_{j+1}$.
Since 
we have  $w^{({i_{j+1}})}\in\sigma_1(w^{({i_{j}})})$ ($1\le j<p$), by induction over $j$ we obtain
$w'\in\sigma_1^*(w)$ thus, according to Lemma \ref{A}, $w'\in\sigma_k^*(w)$. 
$\Box$\\~\ \\
\medskip
In view of proving Lemma \ref{C} and Proposition \ref{card-Sk*}, we need some new lemma:
\begin{lemma}
\label{B}
Assume $|A|=2$. For every $w\in A^{\ge k+1}$, 
we have $\sigma_2(w)\subseteq \sigma_k^2(w)$.
\end{lemma}
{\it Proof.}
Set $A=\{0,1\}$. It follows from $\sigma_2\subseteq\sigma_1^2$ that the result holds for $k=1$.
Assume $k\ge 2$ and let $n=|w|$, $w'\in \sigma_2(w)$. 
By construction, there are distinct integers $i_0,j_0\in [1,n]$  such that the following holds:

(a) $w'_i=\overline{w_i}$ if, and only if, $i\in\{i_0,j_0\}$.\\
Since some $(k-1)$-element set $I\subseteq [1,n]\setminus \{i_0,j_0\}$ exists,
words $w'', w'''\in A^n$ such that: 

(b) $w''_{i}=\overline{w_{i}}$ if, and only if, $i\in \{i_0\}\cup I$, and:

(c) $w'''_{i}=\overline{w''_i}$ if, and only if, $i\in\{j_0\}\cup I$.\\
By construction, we have $w''\in\sigma_k(w)$ and $w'''\in\sigma_k(w'')$, thus $w'''\in\sigma_k^2(w)$.
Moreover, the fact that we have $w'''=w'$ is attested by the following equations:

(d) $w'''_{j_0}=\overline{w''_{j_0}}=\overline{w_{j_0}}=w'_{j_0}$,

(e) $w'''_{i_0}=w''_{i_0}=\overline{w_{i_0}}=w'_{i_0}$, and:

(f) for $i\notin \{i_0,j_0\}$: $w'''_{i}=\overline{w''_{i}}=w_i=w'_i$ if, and only if, $i\in I$.
$\Box$\\~\ \\
{\it Proof of Lemma \ref{C}.}
Assume $k$ even.
According to Property (\ref{P})  we have $w'=w\oplus u$ with $|u|_1=k$.
According to (\ref{Q}), $|w'|_1-|w|_1$ is  even: hence (i) follows.
Conversely, assume $|w'|_1-|w|_1$ even and let $u=w\oplus w'$. According to (\ref{Q}), $|u|_1$ is also even, moreover 
according to (\ref{P}) we obtain $w'=\sigma_{|u|_1}(w)$: this implies  $w'\in\sigma_2^*(w)$.  According to Lemma \ref{B},  we have $w'\in\sigma_k^*(w)$: this establishes (ii).
$\Box\\$\ \\
{\it Proof of Proposition \ref{card-Sk*}.}
Let $w\in A^{\ge k}$ and $n=|w|$. (iii) is trivial and  (i) follows from  Lemma \ref{C}(i): indeed, since $k$ is even,  $\sigma_k^*(w)$ is the set of the words  $w'\in A^n$ such that $|w'|_1- |w|_1$ is even.
Assume $k$ odd and let $w'\in A^n\setminus\{w\}$; we will prove that $w'\in\sigma_k^*(w)$.  If $|w'|_1-|w|_1$ is  even, the result comes from  Lemma \ref{C}(ii). Assume $|w'|_1-|w|_1$ odd and let $t\in\sigma_1(w')$, thus $w'\in\sigma_1(t)\subseteq\sigma_{k}\circ\sigma_{k-1}(t)$ that is, $w'\in\sigma_k(t')$ for some $t'\in\sigma_{k-1}(t)$. It follows from $w'\in\sigma_1(t)$ that $|t|_1-|w|'_1$ is odd, whence $|t|_1-|w|_1=(|t|_1-|w'|_1)+(|w'|_1-|w|_1)$ is even: according to Lemma \ref{C}(ii), this implies $t\in\sigma^*_{k}(w)$. But since $k-1$ is even, we have $t'\in\sigma_{k-1}(t)\subseteq\sigma_2^*(t)$: according to  Lemma \ref{B}, this implies $t'\in\sigma_{k}^*(t)$ (we have $|t|=|w'|=n$). We obtain $w'\in\sigma_k(t')\subseteq\sigma_{k}^*(t)\subseteq \sigma_k^*(\sigma_k^*(w))=\sigma_k^*(w)$: this completes the proof.
$\Box$\\~\ \\
\subsection{The consequences for $\sigma$-closed codes}
\label{Cons}
Given a $\sigma_k$-closed code $X\subseteq A^*$, we say that the tuple $(k,A,X)$ satisfies Condition (\ref{D}) if each of the three following properties holds:
\begin{eqnarray}
\label{D} 
{\rm (a)}~k~{\it is~even,}~~~{\rm (b)}~ |A|=2, ~~~{\rm (c)}~X\not\subseteq A^{\le k}.
\end{eqnarray}
We start by proving the following technical result:
\begin{lemma}
\label{v-sk*w}
Assume  $|A|=2$ and $k$ even.  Given a pair of words  $v,w\in A^+$, 
if  $|w|\ge \max\{|v|+1,k+1\}$ then  the set $\sigma_k^*(w)\cup \{v\}$ cannot be  a code.
\end{lemma}
{\it Proof.}
Let $v,w\in A^+$, and $n=|w|\ge \max\{|v|+1,k+1\}$ (hence we have $v\notin\sigma_k^*(w)\subseteq A^{|w|}$).
By contradiction, we assume that $\sigma_k^*(w)\cup \{v\}$ is  a code.
We are in Condition (i) of Proposition \ref{card-Sk*} that is, we have  $\sigma_k^*(w)\in\{A_0^n,A_1^n\}$.
On a first hand, since  $A^{n-1}$ is a right-complete prefix code \cite[Theorem 3.3.8]{BPR10}, it follows from $|v|\le n-1$ that a (perhaps empty) word $s$ exists such that $vs\in A^{n-1}$. On another hand, it  follows from $A^{n-1}A=A^n=A_0^n\cup A_1^n$ that, for each $u\in A^{n-1}$, a unique pair of letters $a_0,a_1$, exists such that $ua_0\in A_0^n$, $ua_1\in A_1^n$ with $a_1=\overline {a_0}$ that is,
$a\in A$ exists with $vsa\in \sigma_k^*(w)$.
According to Lemma \ref{C}(i), $|sav|_1-|w|_1=|vsa|_1-|w|_1$ is even; according to Lemma \ref{C}(ii), this implies $sav\in\sigma_k^*(w)$. Since we have   $(vsa)v=v(sav)$, the set $s_k^*(w)\cup \{v\}$ cannot be a code.
$\Box$\\~\ \\
As a consequence of Lemma \ref{v-sk*w}, we obtain the following result:
\begin{lemma}
\label{sigma**-code}
Given a 
$\sigma_k$-closed code $X\subseteq A^*$,
if $(k,A,X)$ satisfies Condition (\ref{D}) then  either we have  $X\subseteq A^{\le k}$, or we have  $X\in\{A_0^n,A_1^n,A^n\}$ for some $n\ge k+1$.
\end{lemma}
{\it Proof.}
Assume that we have $X\not\subseteq A^{\le k}$. Firstly, consider two words $v, w\in X\cap A^{\ge k+1}$ and
by contradiction, assume $|v|\neq |w|$ that is,  
without loss of generality $|v|+1\le |w|$.
Since $X$ is $\sigma_k$-closed, we have $\sigma_k^*(w)\subseteq X$, whence the set  $\sigma_k^*(w)\cup \{v\}$, which a subset of $X$  is a code: this contradicts the result of Lemma \ref{v-sk*w}.
Consequently, we have $X\subseteq A^{\le k}\cup A^n$, with $n=|v|=|w|\ge k+1$.
Secondly, once more by contradiction assume that   words $v\in  X\cap A^{\le k}$, $w\in X\cap A^{\ge k+1}$ exist.
As indicated above, since $X$ is $\sigma_k$-closed, $\sigma_k^*(w)\cup\{v\}$ is a code:
since we have $|w|\ge k+1$ and $|w|\ge |v|+1$, once more this contradicts the result of Lemma \ref{v-sk*w}.
As a consequence, if  $X\not\subseteq A^{\le k}$ then necessarily we have $X\subseteq A^n$, for some $n\ge k+1$.
With such a condition, according to Proposition \ref{card-Sk*} for each pair of words $v,w\in X$, we have $\sigma_k^*(v)$, $\sigma_k^*(w)\in \{A_0^n,A_1^n\}$: this implies $X\in\{A_0^n,A_1^n,A^n\}$.
$\Box$\\~\ \\
According to Lemma \ref{sigma**-code}, with Condition (\ref{D}) no $\sigma_k$-closed code can simultaneously possess words in $A^{\le k}$
and words in $A^{\ge k+1}$.
\begin{lemma}
\label{Code-sigma-fini}
Given a  $\sigma_k$-closed code $X\subseteq A^*$, if $(k,A,X)$ does not satisfy Condition \ref{D} then
either we have $X\subseteq A^{\le k}$, or  we have $X=A^n$, with  $n\ge k+1$.
\end{lemma}
{\it Proof.}
If Condition (\ref{D}) doesn't hold then exactly  one of the three following conditions holds:

(a)  $X\subseteq A^{\le k}$;

(b) $X\not\subseteq A^k$ and $|A|\ge 3$;

(c)  $X\not\subseteq A^{\le k}$ with $|A|=2$ and $k$ odd.\\
With each of the two last conditions, let $w\in X\cap A^{\ge k+1}$. 
Since $X$ is $\sigma_k$-closed, according to the propositions \ref{A>2-card-Sk*} and \ref{card-Sk*}(ii), we have $A^n=\sigma_k^*(w)\subseteq\sigma_k^*(X)$. 
Since $A^n$ is a maximal code, it follows from Lemma \ref{iteration-tau-sur-X}(iii) that
$X=A^n$. 
$\Box$\\~\ \\
As a consequence, every $\sigma_k$-closed code is finite. In addition, we state:
\begin{theorem}
\label{sigma*-complete}
 Given a  complete $\sigma_k$ ($\Sigma_k$, $ \Lambda_k$)-closed code $X$, exactly one of the following conditions holds:

{\rm (i)}
$X$ is a subset of $A^{\le k}$;

{\rm (ii)}  a unique integer $n\ge k+1$ exists such that $X=A^n$.\\
In addition, every $\Sigma_k$($\Lambda_k$)-closed code is equal to $A^n$, for some $n\ge 1$.
\end{theorem}
{\it Proof.}
Let $X$ be a   complete $\sigma_k$-closed code.  If Condition (\ref{D}) does not hold, the result is expressed by Lemma \ref{Code-sigma-fini}.
Assume that  Condition (\ref{D}) holds with $X\not\subseteq A^{\le k}$. 
According to Lemma \ref{sigma**-code}, in any case some integer $n\geq k+1$ exists such that $X\in\{A_0^n,A_1^n,A^n\}$. 
Taking for $\pi$ the uniform distribution, we have $\pi(A_0^n)=\pi(A_1^n)=1/2$ and $\pi(A^n)=1$ thus, according to Theorem \ref{classic}: $X=A^n$.
Recall that we have $\sigma_1^*(w)=A^{|w|}$ (eg. \cite{K05}).
Let $w\in X$ and $n=|w|$;
 if $X$ is $\Sigma_k$-closed, we have  $A^n= \sigma_1^*(X)\subseteq \Sigma_k^*(X)\subseteq X$  thus $X=A^n$ (indeed, $A^n$ is a maximal code).
Since  $\Sigma_k\subseteq \Lambda_k$,
if $X$ is $\Lambda_k$-closed then it is $\Sigma_k$-closed, thus we have $X=A^n$.
$\Box$\\~\ \\
As a corollary, 
in the family of $\Sigma_k$($\Lambda_k$)-closed codes, maximality and completeness are equivalent notions. With regard to $\sigma_k$-closed codes, things are otherwise: indeed, as shown in \cite{R77}, there are finite codes that have no finite completion. 
Let $X$ be one of them,  and $k=\max\{|x|:x\in X\}$. By definition $X$ is $\sigma_k$-closed. Since every $\sigma_k$-closed code is finite, no complete $\sigma_k$-closed code can contain $X$.
\begin{proposition}
\label{sigma-completion}
Let $X$ be a (finite) non-complete $\sigma_k$-closed code. Then  one can decide whether some complete $\sigma_k$-closed code containing $X$ exists.
More precisely,  there is only a finite number of such codes, each of them being computable, if any.
\end{proposition}
{\it Proof sketch.}
We draw  the scheme of an algorithm that allows to compute every complete $\sigma_k$-closed code ${\hat X}$ containing $X$.
In a first step, we compute $Y=X\cap A^{\le k}$. 
If $Y=X$,  according to Theorem \ref{sigma*-complete}, we have  ${\hat X}\subseteq A^{\le k}$: ${\hat X}$, if any,  can be computed in a finite number of steps.
Otherwise, ${\hat X}$ exists if, and only if, for some $n\ge k+1$ we have $X\subseteq A^n$: this can be straightforwardly checked.
$\Box$
\bibliography{mysmallbib}{}
\bibliographystyle{splncs04}
\end{document}